\pdfoutput=1
\documentclass[11pt]{article}
\usepackage{acl}
\usepackage{times}
\usepackage{latexsym}
\usepackage[T1]{fontenc}
\usepackage[utf8]{inputenc}
\usepackage{microtype}
\IfFileExists{inconsolata.sty}{\usepackage{inconsolata}}{}
\usepackage{amsmath,amssymb}
\usepackage{booktabs}
\usepackage{graphicx}
\usepackage{xcolor}
\usepackage{tikz}
\usetikzlibrary{positioning,arrows.meta,calc}
\newcommand{\mainimage}[1]{%
  \ifdefined\layoutdraft
    \fbox{\parbox[c][0.32\textheight][c]{\dimexpr\linewidth-2\fboxsep-2\fboxrule\relax}{\centering
    Layout placeholder: original figure not supplied.\\\path{#1}}}%
  \else
    \includegraphics[width=\textwidth]{#1}%
  \fi}
\newcommand{\appendiximage}[1]{%
  \ifdefined\layoutdraft
    \fbox{\parbox[c][0.29\textheight][c]{\dimexpr\linewidth-2\fboxsep-2\fboxrule\relax}{\centering
    Layout placeholder: original figure not supplied.\\\path{#1}}}%
  \else
    \includegraphics[width=\textwidth]{#1}%
  \fi}
\title{From Self-Confidence to Statistical Accuracy: Scaling Human-in-the-Loop Evaluation with Direct and LLM-Assisted Auditing}
\author{
\textbf{Harine Choi}\textsuperscript{1},
\textbf{Chenchen Kuai}\textsuperscript{2},
\textbf{Zihao Li}\textsuperscript{3},
\textbf{Jiwan Jiang}\textsuperscript{2} \\
\textbf{Yunlong Zhang}\textsuperscript{2},
\textbf{Zhiwen Fan}\textsuperscript{4},
\textbf{Zhengzhong Tu}\textsuperscript{5},
\textbf{Yang Zhou}\textsuperscript{2}
\\[3pt]
\textsuperscript{1}Texas A\&M Transportation Institute, Texas A\&M University System \\
\textsuperscript{2}Zachry Department of Civil and Environmental Engineering, Texas A\&M University \\
\textsuperscript{3}Department of Civil, Construction and Environmental Engineering, Marquette University \\
\textsuperscript{4}Department of Electrical and Computer Engineering, Texas A\&M University \\
\textsuperscript{5}Department of Computer Science and Engineering, Texas A\&M University
}

\begin{document}
\maketitle

\begin{abstract}
For large language models (LLMs) and multimodal large language models (MLLMs), model-reported self-confidence is often miscalibrated, yet it may still be useful for efficient evaluation. At the same time, LLM-as-a-judge offers a scalable alternative to direct human auditing, but its practical benefit depends on calibration quality and inference cost.

This paper develops a unified statistical framework for both evaluation routes under alternative sampling strategies. For direct auditing, confidence-stratified sampling is used to recover population accuracy from partially verified outputs without requiring calibrated confidence. Finite-sample uncertainty is characterized, and the analysis shows how highly uneven confidence strata affect audit efficiency and label allocation. For judge-assisted evaluation, the judge is modeled through sensitivity and false-positive rate, yielding an error-corrected estimator with calibration uncertainty. The two routes are then compared through human-label and total-cost efficiency boundaries.

Experiments on MMLU-Pro and MMMU-Pro show that miscalibrated self-confidence can still support statistically valid stratified auditing. Under highly imbalanced confidence distributions, proportional allocation substantially reduces estimation error relative to equal-per-stratum allocation. For judge-assisted evaluation, only a small subset of target--verifier pairs satisfies the idealized human-label efficiency condition, and automated inference costs can further limit potential savings. These findings clarify the statistical and economic conditions under which confidence-guided auditing and automated verification are useful.
\end{abstract}

\section{Introduction}
Large language models (LLMs) and multimodal large language models (MLLMs) are increasingly evaluated at scales that make exhaustive human verification impractical~\citep{liang2022holistic, DBLP:conf/acl/LeeKPKS24, DBLP:conf/icml/ChiangZ0ALLZ0JG24, DBLP:conf/icml/ChenCZWLZZ00024}. Since obtaining reliable reference labels can constitute a substantial fraction of evaluation cost~\citep{DBLP:conf/icml/KossenFGR21, DBLP:conf/icml/BoyeauALYMJ25}, only a subset of model outputs may be inspected in practice. Evaluation then becomes a statistical inference problem: population-level accuracy must be recovered from partial verification, together with uncertainty that reflects the limited labeling budget.

Self-confidence can guide which outputs to audit, but its relationship with correctness varies across models and tasks~\citep{DBLP:conf/emnlp/TianMZSRYFM23, DBLP:conf/iclr/XiongHLLFHH24}. Confidence is often miscalibrated, and disproportionately auditing low-confidence outputs can underrepresent errors among confident predictions. Moreover, confidence strata can differ substantially in size, making audit allocation important for estimation precision. Thus, confidence can guide sampling, but population accuracy must be estimated using the sampling design and stratum sizes.

LLM-as-a-judge provides an alternative evaluation route in which a second LLM evaluates outputs at scale~\citep{DBLP:conf/nips/ZhengC00WZL0LXZ23}. Because the judge is imperfect, human labels are still needed to estimate and correct its error rates, and calibration uncertainty propagates into the final accuracy estimate~\cite{DBLP:journals/corr/abs-2511-21140, chen2026efficient}. The evaluation budget can therefore support either direct auditing or judge calibration. Their relative efficiency depends on judge discrimination, calibration requirements, and the costs of automated inference and human review.

\begin{figure*}[t]
    \centering
    \includegraphics[width=1\textwidth]{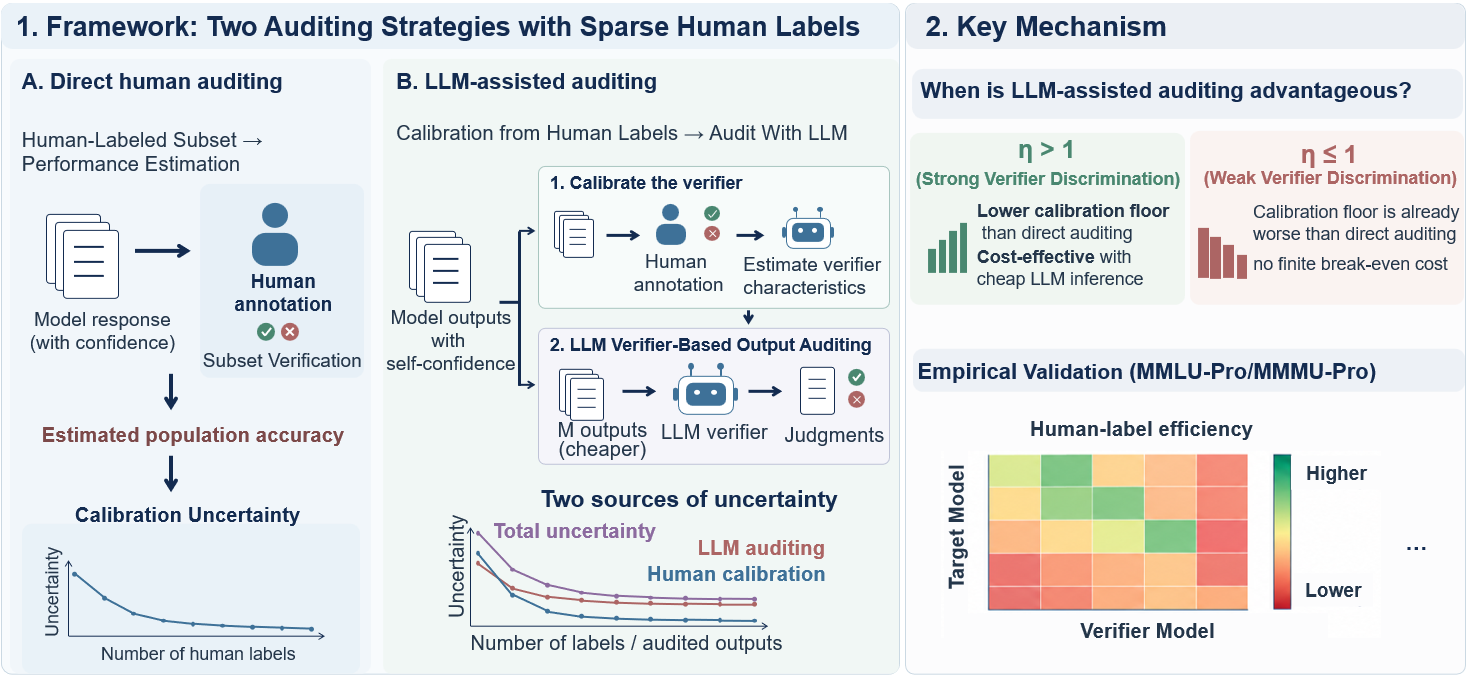}
    \caption{Overview of the proposed evaluation framework}
    \label{fig:overview}
\end{figure*}

In this work, confidence-guided auditing and judge-assisted evaluation are unified within a common statistical framework. For direct auditing, model confidence is used as a stratification variable rather than a trusted correctness probability, enabling population accuracy and finite-sample uncertainty to be estimated from partial verification. For judge-assisted evaluation, the judge is characterized by sensitivity and false-positive rate, yielding an error-corrected estimator with finite calibration uncertainty. The two routes are compared through human-label and cost-efficiency boundaries that identify when automated judging is preferable at the same target precision. Figure~\ref{fig:overview} summarizes the overall framework. We evaluate the approach on MMLU-Pro and MMMU-Pro, covering text and multimodal reasoning.

\section{Related Work}

\paragraph{Confidence, calibration, and selective evaluation.}
Model confidence has long been studied as a signal of predictive uncertainty. Calibration measures whether reported probabilities agree with empirical correctness, with methods such as temperature scaling used to correct systematic errors on labeled validation data~\citep{pmlr-v70-guo17a}. For language models, uncertainty can be elicited directly through verbalized confidence or inferred from model outputs and response consistency~\citep{lin2022teaching, kadavath2022language, DBLP:conf/emnlp/TianMZSRYFM23, DBLP:conf/iclr/KuhnGF23}. Empirical studies nevertheless show substantial miscalibration across models, tasks, and modalities~\citep{DBLP:conf/iclr/XiongHLLFHH24, jangconfidence, DBLP:conf/emnlp/XuanZQWY25}. Selective prediction uses uncertainty to trade coverage for error control~\citep{DBLP:journals/corr/GeifmanE17}. Population evaluation differs because confidence-guided review determines which correctness labels become observable. Estimating overall accuracy from this selected subset therefore raises a verification-bias problem~\citep{begg1983assessment}, requiring the sampling mechanism to be accounted for explicitly.

\paragraph{Label-efficient statistical evaluation.}
Auxiliary model information can improve evaluation through both label acquisition and estimation. Active testing and Active Surrogate Estimators use adaptive sampling and learned surrogates to reduce uncertainty in estimated model performance~\citep{DBLP:conf/icml/KossenFGR21, DBLP:journals/corr/abs-2202-06881}. Prediction-powered inference (PPI) uses a small labeled sample to correct estimates based on abundant model predictions~\cite{angelopoulos2023prediction}, a principle applied to model evaluation by AutoEval~\citep{DBLP:conf/icml/BoyeauALYMJ25}. StratPPI combines this correction with stratified sampling and label allocation to exploit differences in automatic-label quality across subpopulations~\citep{DBLP:conf/nips/FischMHDGC24}. These approaches make both the use of auxiliary predictions and the allocation of labels part of the evaluation design. PPI++ and Adaptive AutoEval adjust reliance on predictions according to their usefulness~\citep{angelopoulos2023ppiplusplus, DBLP:conf/nips/ParkZS25}. With few labels, estimating prediction weights can itself introduce substantial uncertainty, as shown by studies of post-hoc regression and finite-sample PPI++ behavior~\citep{DBLP:conf/icml/EyreM25, DBLP:journals/corr/abs-2505-20178}. Label savings therefore depend on the available budget as well as the auxiliary signal. We examine this design problem for confidence-stratified direct auditing, characterizing estimation precision under fixed allocations and allocations learned from pilot audits.

\paragraph{LLM-as-a-judge and noisy verification.}
LLM-based evaluators have been studied extensively as scalable substitutes for human assessment, including G-Eval, MT-Bench, and learned judge models~\citep{DBLP:conf/emnlp/LiuIXWXZ23, DBLP:conf/nips/ZhengC00WZL0LXZ23, zhu2025judgelm}. However, systematic effects such as position, self-preference, knowledge, and format bias can substantially influence their decisions~\citep{DBLP:conf/nips/PanicksseryBF24}. For binary correctness, judge-assisted evaluation can instead be viewed as prevalence estimation under an imperfect screening test~\citep{rogan1978estimating}. A judge's sensitivity and false-positive rate determine how its acceptance rate maps to target-model accuracy, while finite calibration data introduce additional uncertainty. Recent work has incorporated such calibration into statistical estimation and hypothesis testing~\citep{DBLP:journals/corr/abs-2511-21140, DBLP:conf/iclr/FengSBGR26, chen2026efficient}. Building on this literature, we compare direct auditing and judge-assisted estimation at a common target precision and derive both human-label and total-cost efficiency boundaries.

\section{Statistical Framework and Derivations}
\label{sec:method}

We consider two uses of limited human labels: directly auditing target outputs and calibrating an automated judge. We first establish accuracy estimation and uncertainty for each route, then compare the judge's label and cost requirements against direct random auditing.

\subsection{Population Accuracy and Identification}

Consider a fixed population of $N$ model outputs with reported confidence $C_i$ and binary correctness $Z_i\in\{0,1\}$. Only a subset of the correctness labels is revealed through auditing. Confidence is partitioned into $K$ nonempty strata with sizes $N_j$, population weights $\pi_j=N_j/N$, and stratum accuracies $\theta_j$. The population quantity of interest is
\begin{equation*}
\theta_N=\frac{1}{N}\sum_{i=1}^{N}Z_i
       =\sum_{j=1}^{K}\pi_j\theta_j.
\end{equation*}

Suppose that $n_j\geq1$ outputs are sampled uniformly without replacement from stratum $j$, with audited set $\mathcal A_j$. The stratified estimator is
\begin{equation}
\widehat\theta_{\mathrm{str}}
=\sum_{j=1}^{K}\pi_j\widehat\theta_j,
\quad
\widehat\theta_j
=\frac{1}{n_j}\sum_{i\in\mathcal A_j}Z_i.
\label{eq:estimator}
\end{equation}
Equation~\eqref{eq:estimator} does not require confidence to be calibrated. Confidence is used only to partition the population; statistical validity follows from random sampling within each stratum and population weighting. Miscalibrated confidence can therefore still be useful if it separates outputs with different correctness rates.

Random sampling supports statistical estimation, but the revealed labels alone do not determine every unaudited outcome. To distinguish these two points, let $p_j$ denote the audited fraction of stratum $j$, $\mu_j$ the observed audited accuracy, and $\nu_j$ the unknown accuracy among unaudited outputs. Then
\begin{equation*}
\theta_j=p_j\mu_j+(1-p_j)\nu_j.
\end{equation*}
Without restrictions beyond $0\leq\nu_j\leq1$, the population accuracies compatible with the revealed labels satisfy
\begin{equation*}
\theta_N\in\Bigl[\sum_j\pi_jp_j\mu_j,\,
\sum_j\pi_j\{p_j\mu_j+1-p_j\}\Bigr].
\end{equation*}
These are identification bounds, not sampling confidence intervals. They describe what the observed labels alone establish; the sampling design provides probabilistic guarantees on estimation error, as considered next.

\subsection{Finite-Sample Accuracy and Audit Precision}

For fixed audit allocations and simple random sampling within each stratum,
\begin{equation*}
\mathbb E(\widehat\theta_j)=\theta_j,
\quad
\mathbb E(\widehat\theta_{\mathrm{str}})=\theta_N.
\end{equation*}
For independent stratum samples,
\begin{equation}
\operatorname{Var}(\widehat\theta_{\mathrm{str}})
=
\sum_j\pi_j^2
\left(1-\frac{n_j}{N_j}\right)
\frac{S_j^2}{n_j}.
\label{eq:stratified-variance}
\end{equation}
where $S_j^2$ denotes the finite-population variance of correctness within stratum $j$. Thus, estimation precision depends on stratum weights, within-stratum variability, and audit allocation.

A distribution-free guarantee can also be obtained. Each audited binary observation in stratum $j$ contributes weight $\pi_j/n_j$ to the estimator. Hoeffding's inequality, which also applies to sampling without replacement, gives
\begin{equation}
\Pr\!\left(
|\widehat\theta_{\mathrm{str}}-\theta_N|\geq\epsilon
\right)
\leq2\exp\!\left(
-\frac{2\epsilon^2}{\sum_j\pi_j^2/n_j}
\right).
\label{eq:bound}
\end{equation}
This bound is conservative because it does not exploit finite-population corrections.

For error tolerance $\epsilon$ and failure probability $\delta$, it suffices that
\begin{equation*}
\sum_j\frac{\pi_j^2}{n_j}
\leq\frac{2\epsilon^2}{\log(2/\delta)}.
\end{equation*}
This condition translates a precision target into an audit-budget requirement. Ignoring integer and capacity constraints, proportional allocation minimizes the Hoeffding denominator, whereas Neyman allocation minimizes estimator variance. The two allocations therefore optimize different precision criteria. Appendix~\ref{app:allocation} provides the allocation rules and the corresponding estimator for two-stage adaptive auditing.

\subsection{Accuracy Recovery with an LLM Judge}

Direct auditing obtains human labels for target outputs. An alternative uses human labels to calibrate a second model, which then judges a larger set of outputs. Its decisions must be treated as noisy measurements of correctness.

For a randomly drawn target output, write $\theta=\theta_N$ and let $W\in\{0,1\}$ indicate whether the judge accepts its answer as correct. Define
\begin{equation*}
\begin{aligned}
s&=\Pr(W=1\mid Z=1),\\
\alpha_v&=\Pr(W=1\mid Z=0),
\end{aligned}
\end{equation*}
where $s$ is judge sensitivity and $\alpha_v$ is its false-positive rate. The acceptance probability $q=\Pr(W=1)$ satisfies
\begin{equation}
q=s\theta+\alpha_v(1-\theta),
\quad
\theta=\frac{q-\alpha_v}{s-\alpha_v}.
\label{eq:recovery}
\end{equation}
The discrimination gap is $\Delta_v=s-\alpha_v$. Recovery requires $\Delta_v\neq0$, with $\Delta_v>0$ the intended operating regime. High overall judgment accuracy does not ensure a large gap: a judge that accepts most answers may appear accurate when the target is usually correct, while poorly distinguishing correct from incorrect outputs.

Suppose $s$ is estimated from $n_1$ human-verified correct outputs, $\alpha_v$ from $n_0$ human-verified incorrect outputs, and $q$ from $M$ automated judgments. Assume independent samples, with calibration examples representative of the evaluation population within each correctness class. The corrected estimator is
\begin{equation*}
\widehat\theta_v
=\frac{\widehat q-\widehat\alpha_v}
{\widehat s-\widehat\alpha_v}.
\end{equation*}
Ignoring finite-population corrections, a first-order delta-method approximation gives
\begin{equation}
\begin{aligned}
\operatorname{Var}(\widehat\theta_v)
\approx\frac{1}{\Delta_v^2}\Biggl[
&\frac{q(1-q)}{M}+\frac{\theta^2s(1-s)}{n_1}\\
&+\frac{(1-\theta)^2\alpha_v(1-\alpha_v)}{n_0}
\Biggr].
\end{aligned}
\label{eq:full-var}
\end{equation}
The first term reflects uncertainty in the automated acceptance rate; the other two reflect uncertainty in human calibration. Increasing $M$ reduces the first term but cannot remove calibration uncertainty. A small $\Delta_v$ amplifies all three contributions.

\subsection{Human-Label Budget and Label Efficiency}

We now compare the human labels required to achieve the same target variance $\tau^2$.

For direct random auditing, ignoring finite-population corrections,
\begin{equation*}
B_{\mathrm{direct}}
=\frac{\theta(1-\theta)}{\tau^2}.
\end{equation*}
Confidence-stratified auditing can improve on this baseline when confidence separates groups with different accuracies. The comparison below nevertheless uses direct random auditing as its reference in order to isolate the contribution of verifier calibration relative to the standard unstratified design.

For judge-assisted evaluation, let the calibration budget be
\begin{equation*}
B_v=n_1+n_0.
\end{equation*}
By Eq.~\eqref{eq:full-var}, its calibration variance is minimized when
\begin{equation*}
n_1:n_0
=\theta\sqrt{s(1-s)}:(1-\theta)\sqrt{\alpha_v(1-\alpha_v)}.
\end{equation*}
This is an oracle allocation because it depends on unknown operating rates and requires class-conditional calibration samples. Since correctness is unknown before auditing, obtaining the required correct and incorrect examples may require additional human screening. These screening costs are excluded from the idealized efficiency comparison.

Under this allocation, define
\begin{equation*}
\begin{aligned}
A_v
&=\frac{\bigl[\theta\sqrt{s(1-s)}
+(1-\theta)\sqrt{\alpha_v(1-\alpha_v)}\bigr]^2}{\Delta_v^2},\\
D_v
&=\frac{q(1-q)}{\Delta_v^2}.
\end{aligned}
\end{equation*}
Equation~\eqref{eq:full-var} then becomes
\begin{equation}
\operatorname{Var}(\widehat\theta_v)
\approx\frac{A_v}{B_v}+\frac{D_v}{M}.
\label{eq:optimized-judge-variance}
\end{equation}
Here, $A_v$ determines the calibration contribution and $D_v$ the automated-sampling contribution.

The label-efficiency ratio is therefore
\begin{equation}
\begin{aligned}
\eta
&=\frac{B_{\mathrm{direct}}}{B_v}
=\frac{\theta(1-\theta)}{A_v}\\
&=\frac{\theta(1-\theta)\Delta_v^2}
{\left[
\begin{gathered}
\theta\sqrt{s(1-s)}\\
{}+(1-\theta)\sqrt{\alpha_v(1-\alpha_v)}
\end{gathered}
\right]^2}.
\end{aligned}
\label{eq:efficiency}
\end{equation}
The ratio $\eta$ characterizes oracle human-label efficiency
when automated-sampling uncertainty is negligible
($M\rightarrow\infty$).
When $\eta>1$, verifier calibration requires fewer human
labels than direct random auditing at equal asymptotic
precision under the assumed class-conditional allocation.
For finite $M$, the additional variance contribution
$D_v/M$ reduces this advantage. Thus, $\eta$ represents
an idealized efficiency criterion rather than realized
label savings. An extension to imperfect human verification
is presented in Appendix~\ref{app:imperfect-human}.

\subsection{Cost-Efficiency Boundary}

The preceding comparison neglects automated inference cost. Let $c_H$ denote the cost of one human reference label and $c_L$ the cost of one automated judge decision. Consider the idealized cost
\begin{equation*}
C_{\mathrm{judge}}=c_HB_v+c_LM,
\end{equation*}
excluding additional screening and judge calls on calibration samples.

Under optimal calibration allocation, minimizing this cost subject to
\begin{equation*}
\frac{A_v}{B_v}+\frac{D_v}{M}\leq\tau^2
\end{equation*}
gives
\begin{equation*}
C_{\mathrm{judge}}^{*}
=\frac{
\left(\sqrt{c_HA_v}+\sqrt{c_LD_v}\right)^2
}{\tau^2}.
\end{equation*}
Direct random auditing costs
\begin{equation*}
C_{\mathrm{direct}}
=\frac{c_H\theta(1-\theta)}{\tau^2}.
\end{equation*}
The optimization uses continuous sample sizes and the preceding variance approximation; its derivation is given in Appendix~\ref{app:cost-boundary}.

Judge-assisted evaluation is therefore less expensive under these assumptions when
\begin{equation}
\left(
\sqrt{A_v}+\sqrt{\frac{c_L}{c_H}D_v}
\right)^2
<\theta(1-\theta).
\label{eq:cost-boundary}
\end{equation}
This condition accounts for both calibration uncertainty and automated-evaluation cost. When $c_L/c_H\rightarrow0$, it reduces to
\begin{equation*}
A_v<\theta(1-\theta)
\quad\Longleftrightarrow\quad
\eta>1,
\end{equation*}
recovering the label-efficiency condition in Eq.~\eqref{eq:efficiency}.

\section{Experiments and Results}
\label{sec:experiments}

\subsection{Experimental Setup}

\paragraph{Benchmarks and scope.}
We evaluate whether confidence and cross-model verification can reduce human labeling in model-accuracy audits. MMLU-Pro~\citep{wang2024mmlu} and MMMU-Pro~\citep{yue2025mmmu} provide exact multiple-choice ground truth for text and multimodal evaluation, respectively. We use 8,000 MMLU-Pro questions and a fixed 1,000-question MMMU-Pro subset, covering ten and eight model runs; primary MMLU-Pro comparisons require at least 80\% coverage. Full model and preprocessing details are given in Appendix~\ref{app:protocol}.

\paragraph{Audit simulation.}
Complete benchmark ground truth is used to simulate partial human verification. Outputs are stratified by confidence, and sampling without replacement is repeated 5,000 times across five audit budgets. Only sampled labels are used for estimation; full ground truth is used to assess error and construct oracle baselines.

\paragraph{Verifier evaluation.}
Each of eight models is evaluated against the other seven, yielding 56 directed target--verifier pairs per benchmark. We report judgment accuracy, sensitivity, false-positive rate, discrimination gap, and label efficiency. The primary analysis uses binary judgments; full details are given in Appendix~\ref{app:protocol}.

\begin{figure*}[t]
    \centering
    \includegraphics[width=0.9\textwidth]{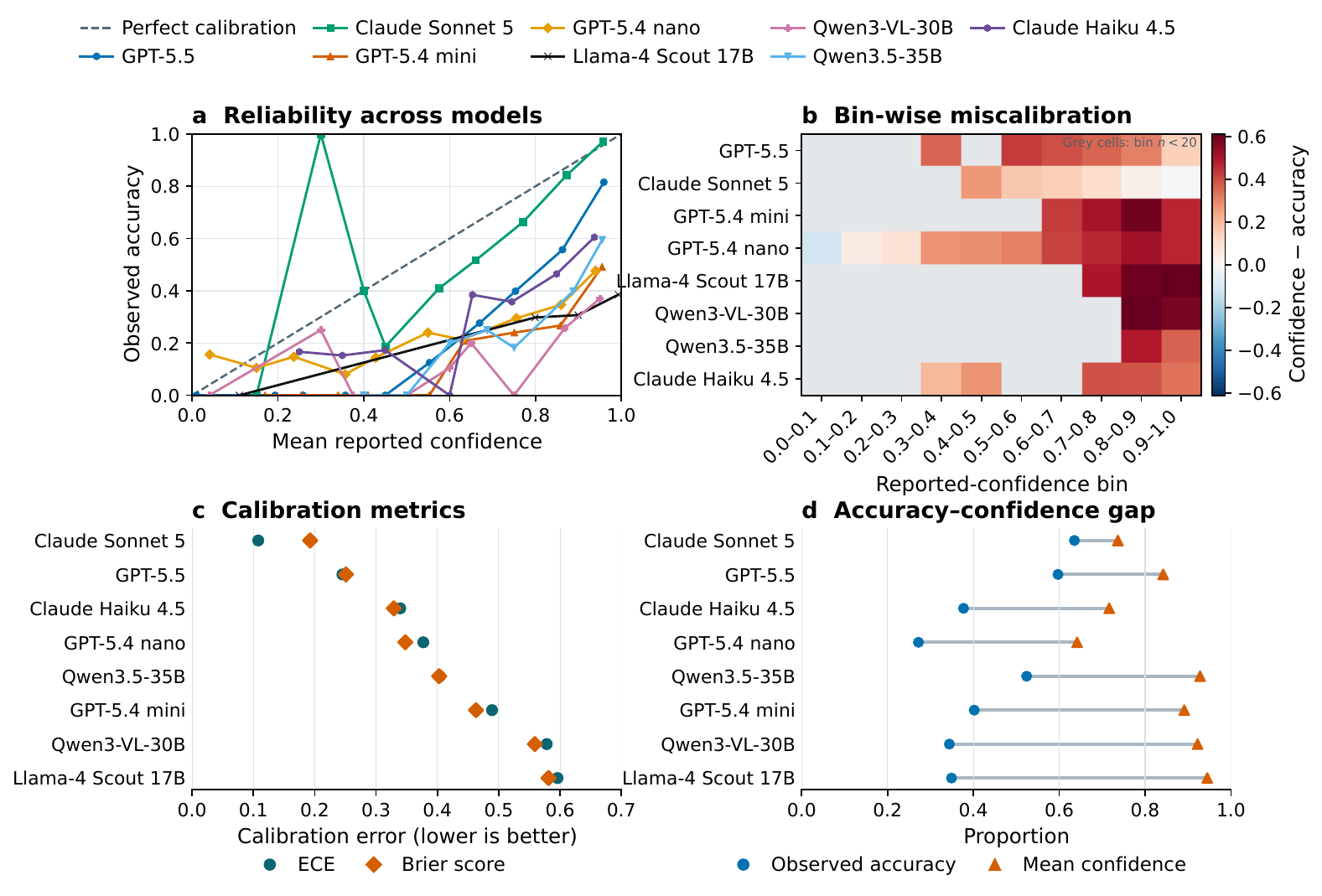}
    \caption{\textbf{MMLU-Pro confidence calibration.}
    \textbf{a}, Reliability curves.
    \textbf{b}, Confidence--accuracy gaps.
    \textbf{c}, Expected calibration error and Brier score.
    \textbf{d}, Accuracy and mean confidence.
    Grey cells contain fewer than 20 observations; daggers mark incomplete runs.}
    \label{fig:mmlu-reliability-calibration}
\end{figure*}

\paragraph{Cost model.}
We estimate each verifier's per-judgment cost from its observed input and output tokens and standard API prices on October 10, 2026. Original answer-generation costs are excluded because they are common to all audit strategies. Estimates also exclude retries, discounts, taxes, and surcharges.

\subsection{Self-Confidence Is Informative but Miscalibrated Across Benchmarks}
Self-reported confidence generally tracks accuracy but is not a calibrated probability of correctness. On MMLU-Pro (Fig.~\ref{fig:mmlu-reliability-calibration}), higher-confidence outputs are typically more accurate, yet panels~\textbf{a} and~\textbf{b} show substantial, model-dependent miscalibration. Claude Sonnet 5 is the best calibrated, while several models remain strongly overconfident; panels~\textbf{c} and~\textbf{d} confirm persistent confidence--accuracy gaps.

The same pattern appears on MMMU-Pro (Appendix Fig.~\ref{fig:mmmu-reliability-calibration}). All eight models are overconfident on average; even Claude Sonnet 5 retains a 10.1-point confidence--accuracy gap and an ECE of 0.108. Across models, accuracy ranges from 27.2\% to 63.5\%, with gaps reaching 59.6 points. Thus, confidence is informative across both benchmarks but unreliable as a direct probability of correctness.

This distinction motivates confidence-stratified auditing. Rather than inferring correctness from confidence, we use it only to partition outputs and estimate population accuracy from representative samples within each stratum.

\begin{figure*}[t]
    \centering
    \includegraphics[width=0.85\textwidth]{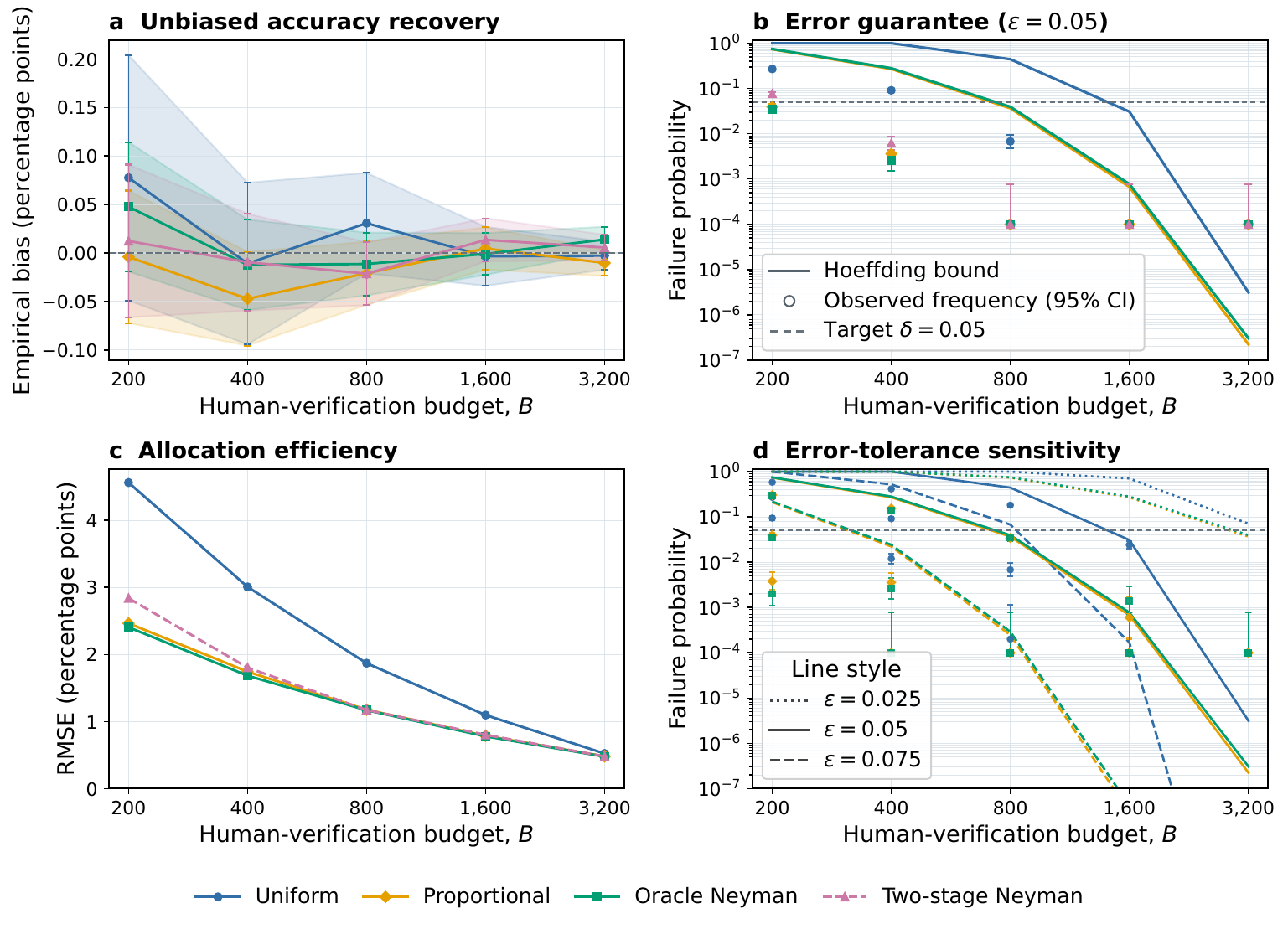}
    \caption{\textbf{Empirical validation of confidence-stratified partial auditing on MMLU-Pro.}
    \textbf{a}, Bias with 95\% Monte Carlo intervals.
    \textbf{b}, Five-point exceedance frequency with 95\% Wilson intervals and Hoeffding bounds.
    \textbf{c}, RMSE.
    \textbf{d}, Exceedance frequencies and bounds.
    Zero observed exceedances are plotted at 0.0001.}
    \label{fig:mmlu-empirical-validation-gpt55}
\end{figure*}

\begin{figure*}[t]
\centering
\includegraphics[width=0.85\textwidth]{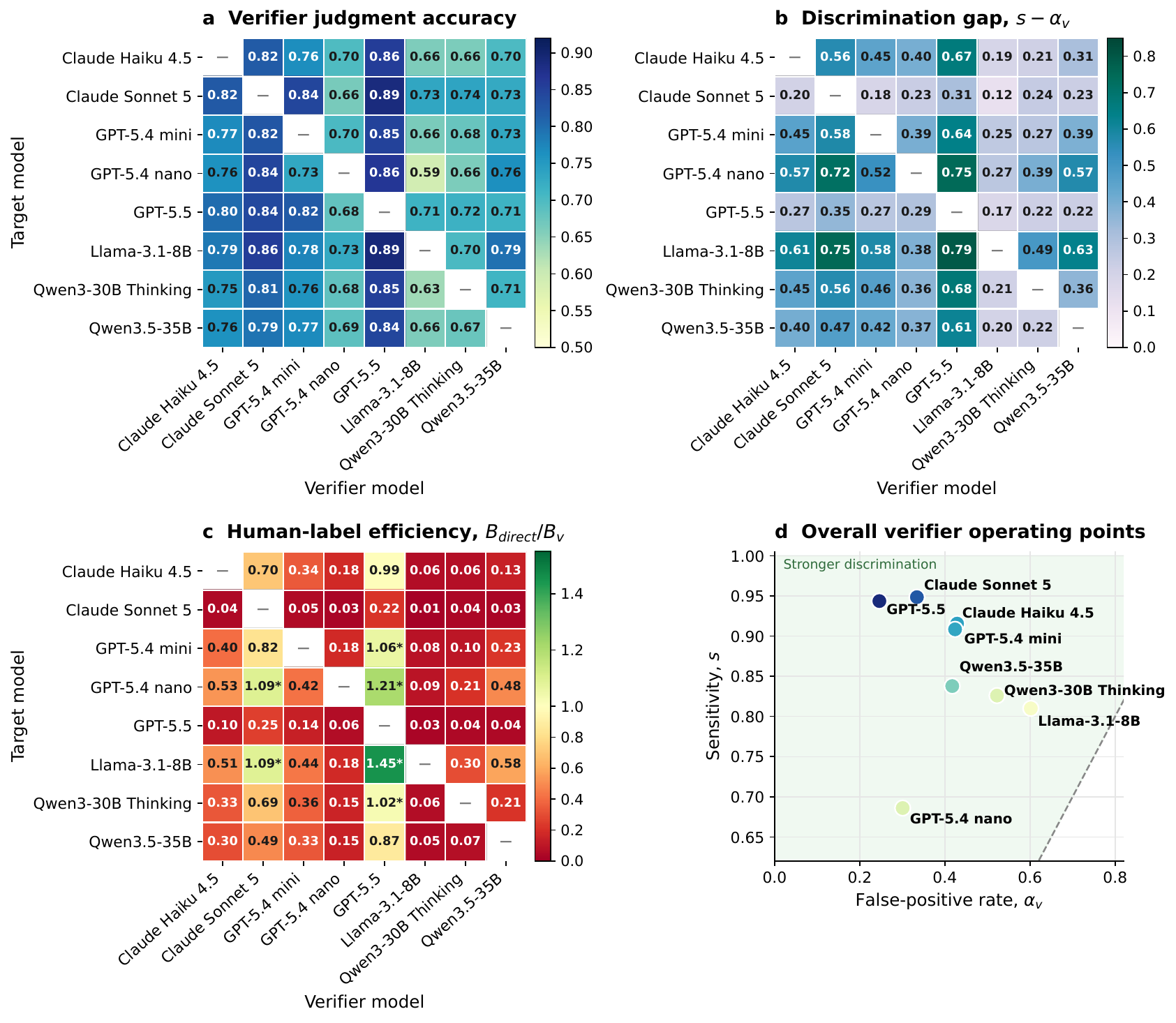}
\caption{\textbf{Observed cross-model verifier performance on MMLU-Pro.}
\textbf{a}, Binary judgment accuracy.
\textbf{b}, Discrimination gap $s-\alpha_v$.
\textbf{c}, Idealized label-efficiency ratio $B_{\mathrm{direct}}/B_v$;
\textbf{d}, Pooled verifier operating points in false-positive-rate--sensitivity space; better discrimination lies closer to the upper-left corner.}
\label{fig:mmlu-verifier-performance}
\end{figure*}

\subsection{Partial Audits Reliably Recover Population Accuracy}

Figure~\ref{fig:mmlu-empirical-validation-gpt55} shows that confidence-stratified auditing reliably recovers GPT-5.5's MMLU-Pro accuracy of 80.54\%. Across budgets and allocation strategies, bias remains near zero (panel~\textbf{a}), while estimation error decreases with additional human labels (panel~\textbf{c}).

Allocation primarily affects precision. Uniform allocation performs worse at small budgets, whereas proportional and oracle Neyman allocations are nearly identical. With 200 labels, RMSE decreases from 4.56 percentage points under uniform allocation to 2.46 and 2.41 percentage points under proportional and oracle Neyman allocation, respectively. This reflects the long-tailed confidence distribution: equal allocation oversamples sparse strata and undersamples dominant ones, while the $\pi_j^2/n_j$ term favors larger strata.

Panels~\textbf{b} and~\textbf{d} show that observed failure probabilities remain below the corresponding Hoeffding bounds. The bounds are conservative but provide valid worst-case guarantees. MMMU-Pro shows the same overall pattern, with greater variability at small budgets (Appendix Fig.~\ref{fig:mmmu-empirical-validation-gpt55}).

\subsection{Verifier Efficiency across Benchmarks}

Figure~\ref{fig:mmlu-verifier-performance} shows that verifier accuracy, discrimination, and label efficiency are distinct. Panel~\textbf{a} reports generally high judgment accuracy, while panel~\textbf{b} shows much larger variation in the discrimination gap $s-\alpha_v$. Panel~\textbf{c} translates these operating characteristics into human-label efficiency: despite strong raw accuracy, only 6 of 56 MMLU-Pro target--verifier pairs achieve $B_{\mathrm{direct}}/B_v>1$. Panel~\textbf{d} explains this pattern: most verifiers have high sensitivity, but several exhibit substantial false-positive rates, weakening discrimination. GPT-5.5 provides the strongest pooled operating point.

\begin{figure}[!htbp]
    \centering
    \includegraphics[width=0.9\columnwidth]
    {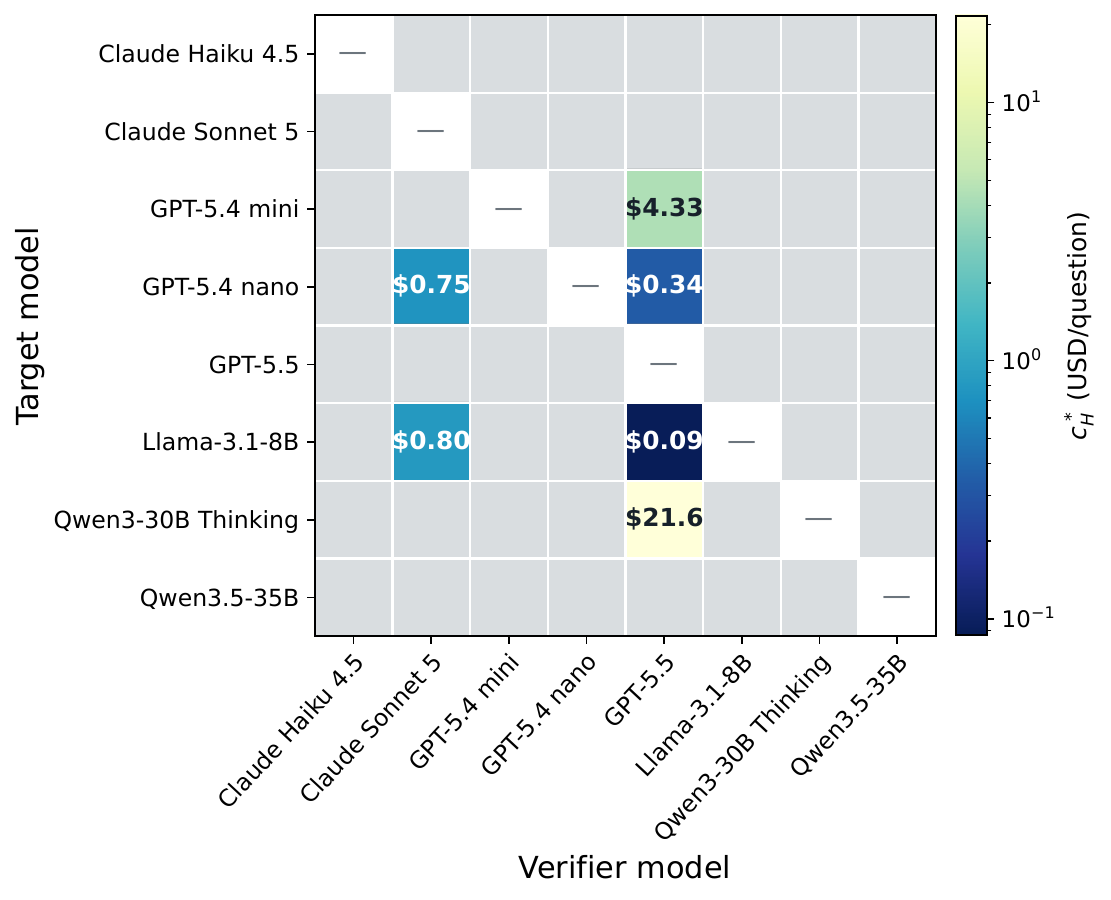}
    \caption{\textbf{Break-even human-label costs on MMLU-Pro.}
    Lower values favor verifier-assisted evaluation. Gray cells have no finite break-even point.}
    \label{fig:verifier-cost-boundary}
\end{figure}

This distinction is important because similar judgment accuracy can imply very different statistical value. For example, Claude Sonnet 5 achieves 0.84 judgment accuracy when verifying GPT-5.5 but only 0.25 label efficiency, whereas GPT-5.5 verifying Llama-3.1-8B achieves 0.89 accuracy and the highest efficiency ratio, 1.45. Thus, efficient population-accuracy recovery depends more directly on separating correct from incorrect answers than on overall agreement.

The pattern is more pronounced on MMMU-Pro
(Appendix Fig.~\ref{fig:mmmu-verifier-performance}). Across 413,562 MMLU-Pro judgments and 55,971 MMMU-Pro judgments, no MMMU-Pro pair exceeds the label-efficiency threshold. GPT-5.5, for example, reaches 0.700 judgment accuracy and 0.844 sensitivity, but its false-positive rate of 0.402 limits the discrimination gap to 0.442.

Figure~\ref{fig:verifier-cost-boundary} adds inference cost. Its six finite break-even cells correspond exactly to the six MMLU-Pro pairs with $B_{\mathrm{direct}}/B_v>1$. The most favorable case is GPT-5.5 verifying Llama-3.1-8B, with $c_H^\star\approx\$0.09$, while the remaining thresholds are approximately \$0.34, \$0.75, \$0.80, \$4.33, and \$21.55. Pairs only slightly above the efficiency boundary can therefore require very high human-label costs before their statistical advantage offsets inference cost.

Overall, verifier discrimination determines the human-calibration burden, while inference cost determines how cheaply automated-audit uncertainty can be reduced. Cheaper LLM inference improves precision by enabling more verifier judgments, but it cannot remove the calibration floor.

\section{Conclusion}
We introduced a unified framework for estimating LLM and MLLM accuracy under limited human verification. Across MMLU-Pro and MMMU-Pro, self-reported confidence was miscalibrated but remained useful for stratified auditing, which recovered population accuracy with low bias. We modeled LLM-based verification through sensitivity and false-positive rate. High judgment accuracy alone does not guarantee efficient auditing; verifier discrimination, calibration uncertainty, and inference cost determine its benefits.

\section*{Limitations}

This study focuses on multiple-choice LLM/MLLM benchmarks with known references and simulated partial auditing, assuming representative sampling and stable verifier behavior. Human-label noise, distribution shift, open-ended evaluation, and continuous verifier confidence are not fully modeled. The cost analysis relies on observed token use and current API-price proxies, while actual human and inference costs may vary; thus, the derived efficiency and break-even boundaries should be interpreted as analytical guidance rather than universal deployment guarantees.

\label{sec:main-end}

\section*{Acknowledgments}

The authors acknowledge support from an NVIDIA Academic Grant Award.

\raggedbottom
\bibliography{custom}

\clearpage
\flushbottom
\appendix

\section{Derivation of the Cost-Efficiency Boundary}
\label{app:cost-boundary}

This appendix derives the minimum cost of verifier-assisted evaluation when both human calibration labels and automated LLM judgments contribute to estimation uncertainty.

From Eq.~\eqref{eq:full-var}, after optimally allocating the human calibration labels between correct and incorrect target outputs, the verifier-assisted variance can be written as
\begin{equation}
\operatorname{Var}(\widehat{\theta}_v)
\approx
\frac{A_v}{B_v}
+
\frac{D_v}{M},
\label{eq:cost-var}
\end{equation}
where
\begin{equation}
A_v=
\frac{
\left[
\theta\sqrt{s(1-s)}
+
(1-\theta)\sqrt{\alpha_v(1-\alpha_v)}
\right]^2
}{
(s-\alpha_v)^2
},
\label{eq:Av}
\end{equation}
and
\begin{equation}
D_v=
\frac{q(1-q)}
{(s-\alpha_v)^2}.
\label{eq:Dv}
\end{equation}
Here, $B_v=n_1+n_0$ is the number of human-labeled examples used to calibrate the verifier, while $M$ is the number of automated verifier judgments. The first term in Eq.~\eqref{eq:cost-var} therefore represents uncertainty due to human calibration, whereas the second represents uncertainty in the estimated automated acceptance rate.

Let $c_H$ denote the cost of one human reference label and $c_L$ the cost of one LLM-judge evaluation. The total cost of verifier-assisted evaluation is
\begin{equation}
C_{\mathrm{judge}}
=
c_H B_v+c_LM.
\label{eq:total-judge-cost}
\end{equation}
For a desired estimation variance $\tau^2$, the minimum-cost design solves
\begin{equation}
\begin{aligned}
\min_{B_v,M>0}\quad
& c_HB_v+c_LM\\
\text{s.t.}\quad
& \frac{A_v}{B_v}+\frac{D_v}{M}
\leq\tau^2.
\end{aligned}
\label{eq:cost-opt}
\end{equation}
At the optimum, the variance constraint is active; otherwise either $B_v$ or $M$ could be reduced while preserving the required precision.

Introducing a Lagrange multiplier $\lambda$, the Lagrangian is
\begin{equation}
\mathcal{L}
=
c_HB_v+c_LM
+
\lambda
\left(
\frac{A_v}{B_v}
+
\frac{D_v}{M}
-
\tau^2
\right).
\end{equation}
The first-order conditions are
\begin{equation}
c_H
=
\lambda\frac{A_v}{B_v^2},
\qquad
c_L
=
\lambda\frac{D_v}{M^2}.
\end{equation}
Eliminating $\lambda$ gives the optimal resource ratio
\begin{equation}
\frac{B_v}{M}
=
\sqrt{
\frac{A_vc_L}
{D_vc_H}
}.
\label{eq:optimal-ratio}
\end{equation}
Thus, more human calibration is favored when calibration uncertainty $A_v$ is large or human labeling is relatively inexpensive, whereas more automated evaluation is favored when acceptance-rate uncertainty $D_v$ is large or automated judgments are inexpensive.

Substituting Eq.~\eqref{eq:optimal-ratio} into the active variance constraint yields
\begin{equation}
M^*
=
\frac{
D_v+\sqrt{A_vD_vc_H/c_L}
}{
\tau^2
},
\label{eq:Mstar}
\end{equation}
and
\begin{equation}
B_v^*
=
\frac{
A_v+\sqrt{A_vD_vc_L/c_H}
}{
\tau^2
}.
\label{eq:Bstar}
\end{equation}
Substitution into Eq.~\eqref{eq:total-judge-cost} gives
\begin{align}
C_{\mathrm{judge}}^*
&=
c_HB_v^*+c_LM^*\\
&=
\frac{
c_HA_v
+
c_LD_v
+
2\sqrt{c_Hc_LA_vD_v}
}{
\tau^2}\\
&=
\frac{
\left(
\sqrt{c_HA_v}
+
\sqrt{c_LD_v}
\right)^2
}{
\tau^2
}.
\label{eq:min-judge-cost}
\end{align}

For comparison, direct random auditing of binary correctness has approximate variance
\[
\operatorname{Var}(\widehat{\theta}_{\mathrm{direct}})
\approx
\frac{\theta(1-\theta)}
{B_{\mathrm{direct}}}.
\]
Achieving the same target variance $\tau^2$ therefore requires
\begin{equation}
B_{\mathrm{direct}}
=
\frac{\theta(1-\theta)}
{\tau^2},
\end{equation}
with corresponding cost
\begin{equation}
C_{\mathrm{direct}}
=
\frac{
c_H\theta(1-\theta)
}{
\tau^2
}.
\label{eq:appendix-direct-cost}
\end{equation}

Verifier-assisted evaluation is more cost efficient whenever
\[
C_{\mathrm{judge}}^*
<
C_{\mathrm{direct}}.
\]
Using Eqs.~\eqref{eq:min-judge-cost} and~\eqref{eq:appendix-direct-cost}, this condition becomes
\begin{equation}
\left(
\sqrt{A_v}
+
\sqrt{
\frac{c_L}{c_H}D_v
}
\right)^2
<
\theta(1-\theta).
\label{eq:appendix-cost-boundary}
\end{equation}
Equation~\eqref{eq:appendix-cost-boundary} is the cost-efficiency boundary reported in the main text. Notably, the target variance $\tau^2$ cancels from the comparison: under the asymptotic approximations used here, changing the desired precision scales both strategies' costs proportionally but does not change which strategy is favored.

For a target--verifier pair satisfying $A_v<\theta(1-\theta)$, solving Eq.~\eqref{eq:appendix-cost-boundary} for the human-label cost gives the break-even value
\begin{equation}
c_H^\star
=
\frac{c_LD_v}
{\left[
\sqrt{\theta(1-\theta)}-\sqrt{A_v}
\right]^2}.
\label{eq:break-even-human-cost}
\end{equation}
Verifier-assisted evaluation has lower optimized total cost than direct auditing when $c_H>c_H^\star$. If $A_v\geq\theta(1-\theta)$, no finite positive human-label cost can make verifier-assisted evaluation preferable under this criterion.

The human-label-only comparison is recovered as a limiting case. When automated judgment cost is negligible relative to human labeling,
$c_L/c_H\rightarrow0$, Eq.~\eqref{eq:appendix-cost-boundary} reduces to
\begin{equation}
A_v<\theta(1-\theta),
\end{equation}
or equivalently
\[
\eta
=
\frac{\theta(1-\theta)}{A_v}
>1.
\]
Thus, the label-efficiency ratio in Eq.~\eqref{eq:efficiency} is the zero-inference-cost limit of the more general cost-efficiency criterion.

These expressions use continuous sample sizes and asymptotic variance approximations. In implementation, $B_v$ and $M$ must be integer-valued and may be constrained by the available evaluation population. Moreover, the correctness classes required to estimate $s$ and $\alpha_v$ are not known before human review, so obtaining $n_1$ correct and $n_0$ incorrect calibration examples may require additional screening. Distribution shift between calibration and evaluation, finite-population corrections, and other operational costs can further affect realized savings. The derived boundary should therefore be interpreted as an analytical cost-efficiency criterion rather than a guaranteed deployment cost.

\section{Evaluation Protocol and Reproducibility}
\label{app:protocol}
\subsection{Model Outputs and Calibration}
For MMLU-Pro, each model received the question and answer options and was asked for JSON containing one option letter, a justification of at most 30 words, and an integer probability of correctness from 1 to 100. Confidence was divided by 100; parsing required a valid option and numeric confidence, with values clipped to the requested range. Hosted runs used structured JSON where supported, a 320-token output limit, and no explicitly supplied temperature. The archived protocol describes vLLM inference with temperature zero, structured JSON, and a 220-token output limit.

The evaluated systems span the GPT family from OpenAI~\citep{openai_gpt55,openai_gpt54_mini_nano}, the Claude family from Anthropic~\citep{anthropic_sonnet5,anthropic_haiku45}, the Gemini family from Google~\citep{google_gemini_models}, the Llama family from Meta~\citep{dubey2024llama3,meta_llama4}, and the Qwen family~\citep{yang2025qwen3,qwen35_modelcard,qwen3vl_modelcard}. Exact evaluated API identifiers or open-weight checkpoints are reported below.

The ten available MMLU-Pro calibration runs are Claude Haiku 4.5, Claude Sonnet 5, GPT-5.4 mini, GPT-5.4 nano, GPT-5.5, Gemini 3.1 Flash-Lite, Gemini 3.5 Flash, Llama-3.1-8B, Qwen3-30B-A3B-Thinking-2507, and Qwen3.5-35B-A3B. The non-self verifier experiment uses the eight non-Gemini models in this list. MMLU-Pro calibration comparisons requiring at least 80\% coverage exclude runs with fewer than 6,400 valid outputs from those panels, while descriptive summaries retain them with a dagger marker.

Reliability curves use ten equal-width confidence bins. Expected calibration error weights each bin's absolute confidence--accuracy gap by its population frequency. The Brier score averages squared differences between confidence and binary correctness. Calibration-gap displays suppress bins with fewer than 20 observations on MMLU-Pro or 25 observations on MMMU-Pro.

The MMMU-Pro analysis uses the first 1,000 examples, in dataset order, from the test split of the \texttt{standard (10 options)} configuration of \texttt{MMMU/MMMU\_Pro}, fixed at revision \texttt{563f3e84bb3b90893083a1f039cfa13077f2302b}. The selected questions contain 1,135 associated images. The eight usable runs are GPT-5.5, Claude Sonnet 5, GPT-5.4 mini, GPT-5.4 nano, Claude Haiku 4.5, Qwen3.5-35B, Qwen3-VL-30B, and Llama-4 Scout 17B-16E-Instruct. Claude Sonnet 5 contains 998 valid outputs, and each other run contains 1,000.

\subsection{Fixed and Adaptive Audits}
The initial strata are $[0,0.2],(0.2,0.4],(0.4,0.6],(0.6,0.8]$, and $(0.8,1]$. Adjacent strata are merged until each retained stratum contains at least 25 outputs. Integer allocations obey $1\leq n_j\leq N_j$ and sum to the budget. Uniform allocation distributes counts as evenly as capacities allow. Proportional and oracle Neyman allocations assign each remaining audit to the stratum giving the greatest decrease in their respective convex objectives, $\sum_j\pi_j^2/n_j$ and $\sum_j\pi_j^2S_j^2/n_j$.

Appendix~\ref{app:allocation} gives the allocation formulas, pilot stabilization, and two-stage estimator. The adaptive design is compared empirically but excluded from fixed-allocation concentration validation.

The plotted MMLU-Pro GPT-5.5 audit simulation uses random seed 20260807. The MMMU-Pro multi-model audit simulation uses master seed 20260807 and NumPy's \texttt{SeedSequence} to generate one child stream per model; the plotted GPT-5.5 population uses child seed 4025595370. Every budget--allocation condition uses 5,000 independent audit replications. Bias is the mean estimate minus population accuracy, and RMSE is the square root of the mean squared estimation error. Bias intervals use the Monte Carlo standard error, whereas exceedance frequencies use Wilson intervals. Error tolerances are 2.5, 5, and 7.5 percentage points. Zero observed exceedances are plotted at 0.0001 only to permit logarithmic axes; their recorded frequency is zero.

Additional MMLU-Pro populations described in the source are Claude Sonnet 5 (7,561 outputs; accuracy 0.872900), GPT-5.4 nano (8,000; 0.407750), and Qwen3.5-35B-A3B (8,000; 0.635125). The present empirical audit figures depict GPT-5.5; no unprovided audit figure for these additional populations is claimed.

\subsection{Verifier Judgments}
A verifier received the question, answer options, and the target model's selected answer and returned a correct/incorrect judgment together with an estimated probability that the selected target answer was correct. The primary analysis uses the binary judgment. On MMLU-Pro, 19 of 413,581 requested non-self judgments were unparseable, leaving 413,562 valid judgments. MMMU-Pro target--verifier pairs contain 997--1,000 usable judgments, yielding 55,971 valid judgments in total. Rates are computed separately for each target--verifier pair; pooled operating points aggregate eligible confusion-matrix counts. The full available pair population is used for the efficiency calculation rather than a selected high-confidence subset.

Sensitivity is the fraction of correct target answers accepted; the false-positive rate is the fraction of incorrect answers accepted. Judgment accuracy counts both correct acceptance and correct rejection. Complete-sample estimates are substituted into Eq.~\eqref{eq:efficiency}. This is a descriptive comparison, with neither a held-out calibration pool nor an implemented sequential class-enrichment policy.

\subsection{Artifact Availability}

MMLU-Pro and MMMU-Pro are publicly available benchmark datasets. Prompts, selected model outputs, cross-model verifier judgments, the MMMU-Pro subset manifest, and derived metric and coverage summaries are available at \url{https://github.com/harinechoi/confidence-auditing-reproduction}, subject to benchmark licenses and model-provider terms.

\section{Audit Allocation Strategies}
\label{app:allocation}

This appendix provides additional details on the audit-allocation strategies used in the experiments. The total audit budget is denoted by $B$, and $n_j$ denotes the number of audited outputs assigned to confidence stratum $j$. All allocations satisfy
\[
\sum_{j=1}^{K} n_j=B,
\qquad
1\leq n_j\leq N_j,
\]
so that every nonempty stratum receives at least one audit and no stratum is sampled beyond its population size.

\subsection{Fixed Audit Allocations}

We consider three fixed allocation strategies: uniform, population-proportional, and oracle Neyman allocation.

\paragraph{Uniform allocation.}
Uniform allocation distributes the audit budget approximately equally across the $K$ confidence strata,
\[
n_j^{\mathrm{uni}}\approx \frac{B}{K}.
\]
This allocation treats all strata equally regardless of their population size or internal variability. It therefore differs from simple random sampling over the full population. In particular, when confidence scores are highly concentrated in one or two strata, equal allocation can oversample small strata and undersample the regions containing most model outputs.

\paragraph{Population-proportional allocation.}
Proportional allocation assigns audits according to stratum prevalence,
\begin{equation}
n_j^{\mathrm{prop}}
=
B\pi_j,
\qquad
\pi_j=\frac{N_j}{N},
\label{eq:prop-allocation}
\end{equation}
before integer and capacity adjustments. This allocation is equivalent, up to stratification constraints, to preserving the population composition in the audited sample.

Its statistical role can be seen from the Hoeffding bound in Eq.~\eqref{eq:bound}, whose exponent depends on
\[
\sum_{j=1}^{K}\frac{\pi_j^2}{n_j}.
\]
For a fixed budget $B$, minimizing this quantity subject to $\sum_jn_j=B$ gives
\[
n_j\propto\pi_j.
\]
Thus, proportional allocation provides the strongest bound within this family of fixed stratified designs. Importantly, this result does not depend on knowing the true stratum accuracies.

\paragraph{Oracle Neyman allocation.}
If the within-stratum correctness variances $S_j^2$ were known, the variance in Eq.~\eqref{eq:stratified-variance} would be minimized by assigning more labels to strata that are both prevalent and internally uncertain. Ignoring integer and capacity constraints, the resulting Neyman allocation is
\begin{equation}
n_j^{\mathrm{Ney}}
=
B
\frac{\pi_jS_j}
{\sum_{k=1}^{K}\pi_kS_k}.
\label{eq:neyman-allocation}
\end{equation}
For binary correctness,
\[
S_j^2
\approx
\theta_j(1-\theta_j),
\]
so the largest allocations are generally assigned to large strata whose accuracies are closer to $0.5$. Strata that are almost entirely correct or incorrect require fewer samples because their within-stratum variance is small.

Oracle Neyman allocation is not directly implementable because the $S_j$ values depend on correctness labels that are unknown before auditing. It is therefore used only as a reference for the best precision achievable when the true within-stratum variances are available.

\subsection{Why Proportional and Neyman Allocation Differ}

The proportional and Neyman designs optimize different statistical objectives. To see this, consider the generic problem
\begin{equation}
\min_{\{n_j\}}
\sum_{j=1}^{K}\frac{a_j}{n_j}
\qquad
\text{subject to}
\qquad
\sum_{j=1}^{K}n_j=B.
\label{eq:generic-allocation}
\end{equation}
The Lagrangian is
\[
\mathcal L
=
\sum_j\frac{a_j}{n_j}
+
\lambda\left(\sum_jn_j-B\right).
\]
Setting the derivative with respect to $n_j$ to zero gives
\[
-\frac{a_j}{n_j^2}+\lambda=0,
\]
and therefore
\[
n_j
=
\frac{B\sqrt{a_j}}
{\sum_k\sqrt{a_k}}.
\]

For the Hoeffding denominator,
$a_j=\pi_j^2$, yielding
\[
n_j\propto\pi_j,
\]
which gives proportional allocation. For the estimator variance,
$a_j=\pi_j^2S_j^2$, yielding
\[
n_j\propto\pi_jS_j,
\]
which gives Neyman allocation.

The distinction is therefore fundamental: proportional allocation minimizes a distribution-free concentration criterion, whereas Neyman allocation minimizes variance by exploiting information about within-stratum correctness variability. The latter can outperform proportional allocation when variability differs substantially across strata, but that information must first be estimated.

\subsection{Two-Stage Neyman Auditing}

To approximate Neyman allocation without access to the true $S_j$, we use a two-stage design. In the first stage, a small pilot sample of size $m_j$ is drawn uniformly without replacement from every stratum. In the experiments, $m_j=10$ whenever the available stratum size permits. Let
\[
X_{1j}
=
\sum_{i\in\mathcal A_{1j}}Z_i
\]
denote the number of correct outputs observed in the pilot sample.

The pilot accuracy is stabilized as
\begin{equation}
\widetilde{\theta}_{1j}
=
\frac{X_{1j}+1/2}{m_j+1}.
\label{eq:pilot-rate}
\end{equation}
The small pseudo-count prevents estimated variances from becoming exactly zero when all pilot observations are correct or all are incorrect. The estimated Bernoulli standard deviation is then
\[
\widehat S_j
=
\sqrt{
\widetilde{\theta}_{1j}
\left(1-\widetilde{\theta}_{1j}\right)
}.
\]

After the pilot stage, the remaining audit budget is
\[
B_2
=
B-\sum_jm_j.
\]
The second-stage samples are allocated approximately according to
\[
n_{2j}
\propto
(N_j-m_j)\widehat S_j,
\]
subject to the remaining stratum capacities and the requirement that each nonempty remainder receive sufficient representation. Thus, the pilot is used only to determine where the remaining labels should be spent.

\subsection{Estimation after Adaptive Allocation}

Because the second-stage allocation depends on the pilot outcomes, simply averaging all audited observations within a stratum can introduce bias. The final estimator therefore treats the pilot and the remaining population explicitly.

Let $\widehat\theta_{2j}$ denote the mean correctness among the $n_{2j}$ outputs sampled uniformly from the $N_j-m_j$ outputs not observed in the pilot. The stratum total is estimated by combining the observed pilot total with an expansion estimate of the unaudited remainder:
\begin{equation}
\widehat\theta_{j,\mathrm{2st}}
=
\frac{
X_{1j}
+
(N_j-m_j)\widehat\theta_{2j}
}{
N_j
}.
\label{eq:two-stage}
\end{equation}
The overall population estimator is
\[
\widehat\theta_{\mathrm{2st}}
=
\sum_{j=1}^{K}
\pi_j\widehat\theta_{j,\mathrm{2st}}.
\]

To see why this remains unbiased, condition on the realized pilot sample. Given the pilot, the second-stage allocation $n_{2j}$ is fixed, while the second-stage sample remains uniform over the unsampled portion of the stratum. Therefore,
\[
\mathbb E
\left[
\widehat\theta_{2j}
\mid
\mathcal A_{1j}
\right]
=
\theta_{j,\mathrm{rem}},
\]
where $\theta_{j,\mathrm{rem}}$ is the true accuracy among the $N_j-m_j$ remaining outputs. Hence,
\[
\mathbb E
\left[
X_{1j}
+
(N_j-m_j)\widehat\theta_{2j}
\mid
\mathcal A_{1j}
\right]
=
X_{1j}
+
(N_j-m_j)\theta_{j,\mathrm{rem}},
\]
which equals the true total number of correct outputs in stratum $j$. Consequently,
\[
\mathbb E
\left[
\widehat\theta_{j,\mathrm{2st}}
\right]
=
\theta_j,
\]
and population weighting preserves unbiasedness.

This distinction is important because the pilot observations influence the second-stage sample sizes. A naive pooled mean would treat pilot and second-stage observations as though they had identical inclusion mechanisms, whereas Eq.~\eqref{eq:two-stage} explicitly accounts for the adaptive design.

\section{Extension to Imperfect Human Verification}
\label{app:imperfect-human}

The main analysis treats human reference labels as correct observations of the latent correctness variable $Z$. In practice, human verification may itself be imperfect, particularly for difficult or multimodal questions. This section shows how the framework extends when human labels are noisy.

\subsection{Noisy Human Labels}

Let $H\in\{0,1\}$ denote the human assessment of whether a model answer is correct. Define the human sensitivity and false-positive rate as
\begin{equation*}
\begin{aligned}
s_H&=\Pr(H=1\mid Z=1),\\
\alpha_H&=\Pr(H=1\mid Z=0).
\end{aligned}
\end{equation*}
and let
\begin{equation*}
\Delta_H=s_H-\alpha_H
\end{equation*}
denote human discrimination. Perfect human verification corresponds to $s_H=1$ and $\alpha_H=0$.

For a population with true model accuracy $\theta$, the observed fraction of answers marked correct by humans is
\begin{equation}
r
=
s_H\theta+\alpha_H(1-\theta).
\label{eq:human-noisy-rate}
\end{equation}
Thus, treating imperfect human labels directly as ground truth estimates $r$ rather than $\theta$. The resulting bias is
\begin{equation*}
r-\theta
=
\alpha_H(1-\theta)
-
(1-s_H)\theta.
\end{equation*}
Even small human error rates can therefore produce systematic bias when the evaluation population is large.

If $s_H$ and $\alpha_H$ are known or estimated independently, Eq.~\eqref{eq:human-noisy-rate} can be inverted:
\begin{equation}
\theta
=
\frac{r-\alpha_H}{s_H-\alpha_H}
=
\frac{r-\alpha_H}{\Delta_H}.
\label{eq:human-correction}
\end{equation}
As with an automated verifier, identification requires $\Delta_H\neq0$.

\subsection{Confidence-Stratified Auditing with Imperfect Humans}

For confidence stratum $j$, let
\begin{equation*}
r_j=\Pr(H=1\mid j).
\end{equation*}
If human operating characteristics are stable across confidence strata,
\begin{equation*}
r_j
=
s_H\theta_j+\alpha_H(1-\theta_j),
\end{equation*}
and the corrected stratum accuracy is
\begin{equation}
\theta_j
=
\frac{r_j-\alpha_H}{\Delta_H}.
\label{eq:human-stratum-correction}
\end{equation}

Let
\begin{equation*}
\widehat r_j
=
\frac{1}{n_j}
\sum_{i\in\mathcal A_j}H_i.
\end{equation*}
A corrected population estimator is then
\begin{equation}
\widehat\theta_{\mathrm{str},H}
=
\sum_{j=1}^{K}
\pi_j
\frac{\widehat r_j-\alpha_H}{\Delta_H}.
\label{eq:human-corrected-stratified}
\end{equation}
When $s_H$ and $\alpha_H$ are known and human errors satisfy the assumed measurement model, this estimator remains unbiased for population accuracy. Hence, imperfect human verification does not invalidate confidence-stratified auditing, but it requires an additional measurement-error correction.

With known human operating characteristics, its sampling variance is approximately
\begin{equation}
\operatorname{Var}
(\widehat\theta_{\mathrm{str},H})
=
\frac{1}{\Delta_H^2}
\sum_j
\pi_j^2
\left(1-\frac{n_j}{N_j}\right)
\frac{S_{H,j}^2}{n_j},
\label{eq:human-stratified-var}
\end{equation}
where $S_{H,j}^2$ is the finite-population variance of the human labels in stratum $j$. Human disagreement therefore inflates uncertainty by approximately $1/\Delta_H^2$.

\subsection{Uncertainty in Human Reliability}

In practice, $s_H$ and $\alpha_H$ must themselves be estimated. Suppose an adjudicated calibration set contains $m_1$ truly correct and $m_0$ truly incorrect outputs, from which
$\widehat s_H$ and $\widehat\alpha_H$ are estimated. Let
\begin{equation*}
\widehat r_{\mathrm{str}}
=
\sum_j\pi_j\widehat r_j.
\end{equation*}
The corrected estimator becomes
\begin{equation*}
\widehat\theta_H
=
\frac{
\widehat r_{\mathrm{str}}
-
\widehat\alpha_H
}{
\widehat s_H-\widehat\alpha_H
}.
\end{equation*}

A first-order delta-method approximation gives
\begin{equation}
\begin{aligned}
\operatorname{Var}(\widehat\theta_H)
\approx
\frac{1}{\Delta_H^2}
\Bigg[
&\operatorname{Var}(\widehat r_{\mathrm{str}})
+
\frac{\theta^2s_H(1-s_H)}{m_1}\\
&+
\frac{(1-\theta)^2\alpha_H(1-\alpha_H)}{m_0}
\Bigg].
\end{aligned}
\label{eq:human-full-var}
\end{equation}
The first term is uncertainty from the audit sample, while the remaining terms arise from estimating human reliability. Increasing the audit budget cannot eliminate the latter. Thus, sufficiently precise population evaluation may eventually be limited by uncertainty in the human reference process rather than by the number of audited model outputs.

\subsection{Implications for LLM-as-a-Judge}

Imperfect human labels also affect judge calibration. If the observed human label $H$ is naively treated as true correctness $Z$, the estimated judge sensitivity and false-positive rate characterize agreement with the human annotator rather than agreement with latent correctness. The efficiency boundary in Eq.~\eqref{eq:efficiency} can consequently be distorted.

Under the additional conditional-independence assumption
\begin{equation*}
W\perp H\mid Z,
\end{equation*}
known human error rates permit correction. For example,
\begin{equation*}
\begin{aligned}
&\begin{bmatrix}
\Pr(W=1,H=1)\\
\Pr(W=1,H=0)
\end{bmatrix}\\
&\quad=
\begin{bmatrix}
s_H & \alpha_H\\
1-s_H & 1-\alpha_H
\end{bmatrix}
\begin{bmatrix}
s\theta\\
\alpha_v(1-\theta)
\end{bmatrix}.
\end{aligned}
\end{equation*}
When $\Delta_H\neq0$, the measurement matrix is invertible, allowing the latent quantities $s\theta$ and $\alpha_v(1-\theta)$---and hence the judge operating rates---to be recovered after correcting for human error.

This extension reveals an additional cost of judge-assisted evaluation: human labels used to calibrate the judge must themselves be sufficiently reliable or independently calibrated. Human error therefore does not simply add noise to the evaluation; it introduces another measurement layer whose uncertainty propagates into both direct auditing and automated-judge calibration.

\subsection{Practical Interpretation}

Three regimes are distinguished. If human error is negligible, the main-text analysis applies directly. If human sensitivity and false-positive rates can be estimated from adjudicated examples, Eqs.~\eqref{eq:human-correction}--\eqref{eq:human-full-var} provide an error-corrected extension. If human reliability is unknown, however, population accuracy is generally not point identified from a single noisy human label per output without additional assumptions or repeated/adjudicated annotations.

Accordingly, the perfect-human assumption in the main analysis should be interpreted as defining the statistical value of a trusted reference label. In practical deployments, disagreement resolution, repeated annotation, or expert adjudication can be used to approximate this reference process, with their additional costs incorporated into $c_H$.

\section{Additional Empirical Figures}
\label{app:empirical-figures}
The additional figures report confidence calibration, repeated-audit validation, and cross-model verification on MMMU-Pro.

\begin{figure*}[htbp]
    \centering
    \appendiximage{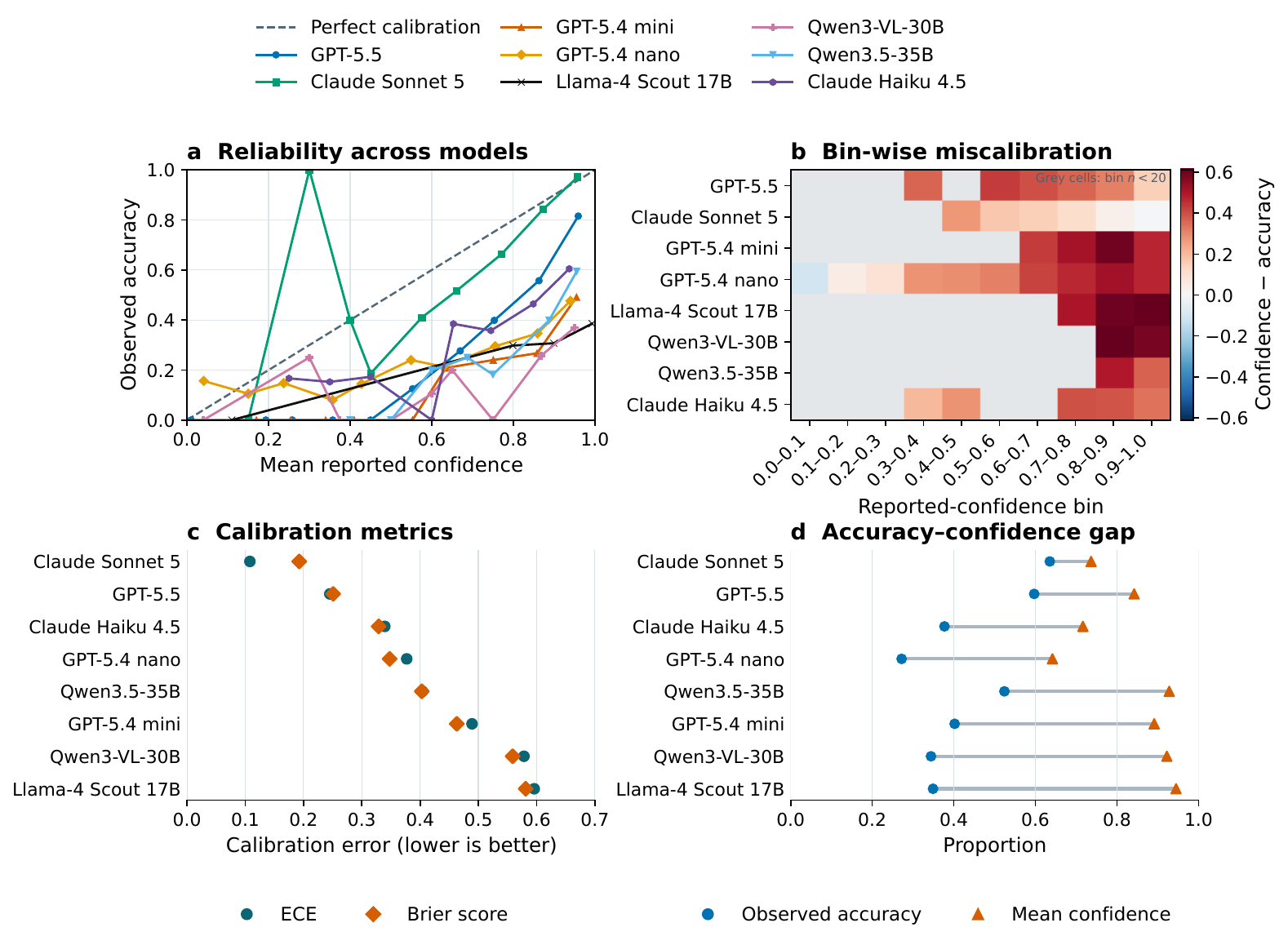}
    \caption{\textbf{MMMU-Pro calibration.} \textbf{a}, Reliability curves. \textbf{b}, Confidence minus accuracy by bin; grey cells have fewer than 25 observations. \textbf{c}, ECE and Brier score. \textbf{d}, Accuracy and mean confidence. Seven runs contain 1,000 outputs; Sonnet 5 contains 998. The diagonal denotes perfect calibration.}
    \label{fig:mmmu-reliability-calibration}
\end{figure*}

\begin{figure*}[htbp]
    \centering
    \appendiximage{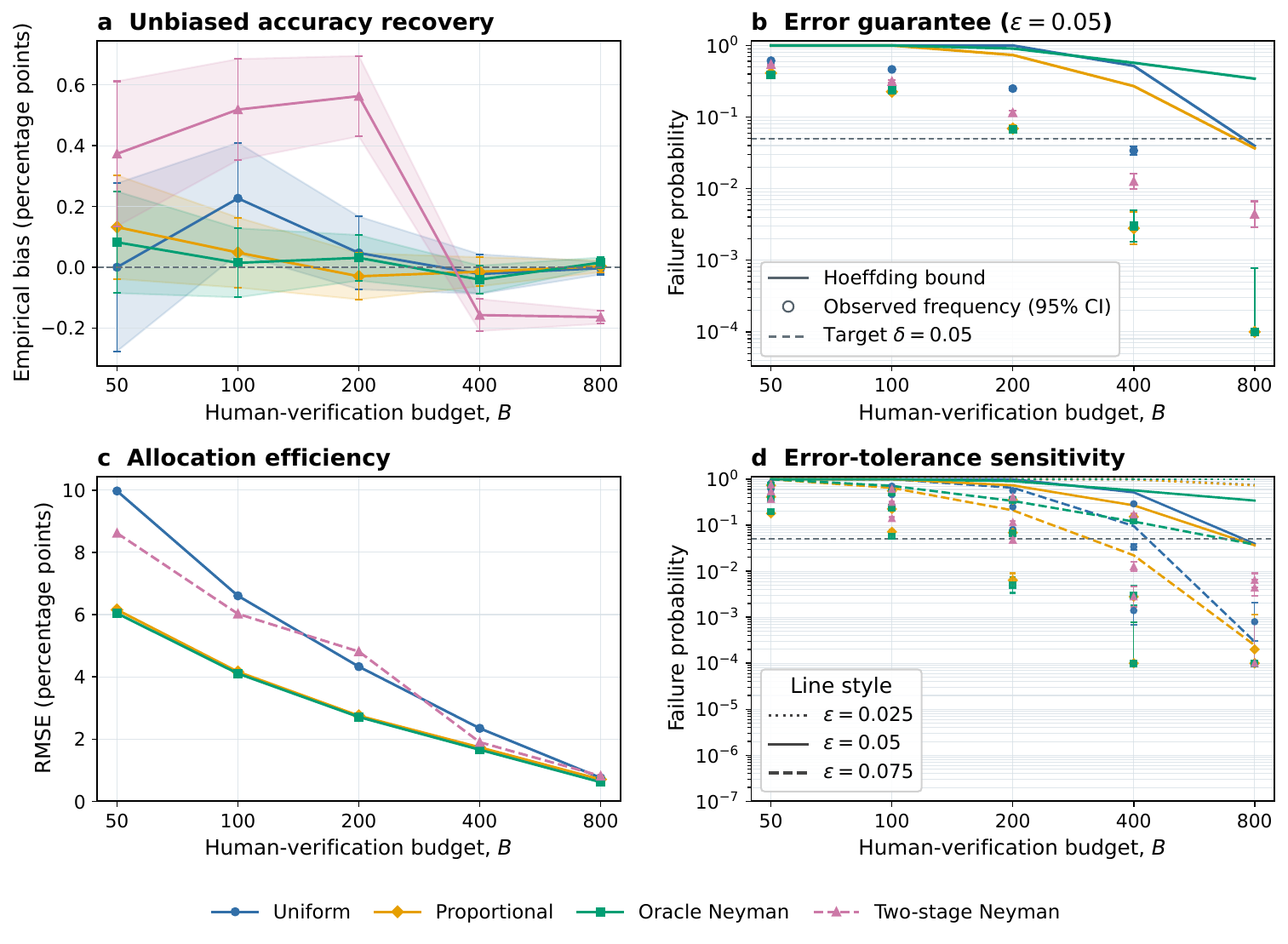}
    \caption{\textbf{MMMU-Pro audits.} GPT-5.5 results over 5,000 replications with four merged strata. Panels follow Fig.~\ref{fig:mmlu-empirical-validation-gpt55}: \textbf{a}, bias; \textbf{b}, five-point exceedances and fixed-allocation bounds; \textbf{c}, RMSE; \textbf{d}, tolerance sensitivity. Intervals are 95\% Monte Carlo (a) and Wilson (b); zero failures are plotted at 0.0001.}
    \label{fig:mmmu-empirical-validation-gpt55}
\end{figure*}

\begin{figure*}[htbp]
\centering
\mainimage{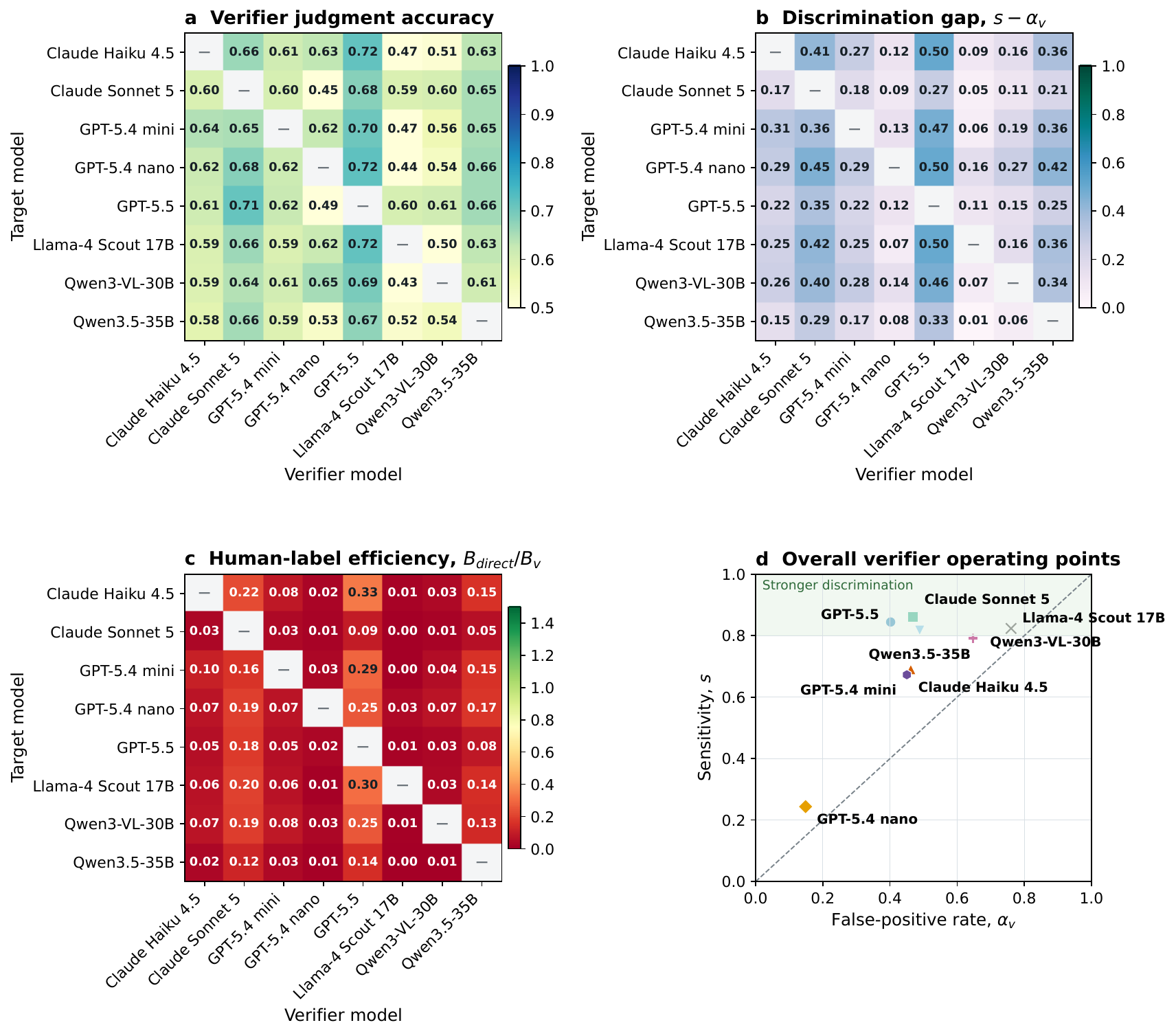}
\caption{\textbf{MMMU-Pro verification.} Panels follow Fig.~\ref{fig:mmlu-verifier-performance}. No pair has an idealized label-efficiency ratio above one. Self-verification is excluded.}
\label{fig:mmmu-verifier-performance}
\end{figure*}

\end{document}